\documentclass[pdflatex,sn-mathphys-ay]{sn-jnl}

\usepackage{graphicx}%
\usepackage{multirow}%
\usepackage{amsmath,amssymb,amsfonts}%
\usepackage{amsthm}%
\usepackage{mathrsfs}%
\usepackage[title]{appendix}%
\usepackage{xcolor}%
\usepackage{textcomp}%
\usepackage{manyfoot}%
\usepackage{booktabs}%
\usepackage{algorithm}%
\usepackage{algorithmicx}%
\usepackage{algpseudocode}%
\usepackage{listings}%
\usepackage{placeins}
\theoremstyle{thmstyleone}%
\theoremstyle{thmstyletwo}%

\theoremstyle{thmstylethree}%

\begin{document}

\title[Article Title]{SIINR: Structurally Informed Implicit Neural Representations for super-resolution with uncertainty quantification of clinical quality diffusion MRI datasets. }


\author*[1]{\fnm{Tom} \sur{Hendriks}}\email{t.hendriks@tue.nl}

\author[2]{\fnm{William} \sur{Consagra}}
\author[1]{\fnm{Anna} \sur{Vilanova}}

\author[3]{\fnm{Yogesh} \sur{Rathi}}
\equalcont{These authors contributed equally to this work.}
\author[1]{\fnm{Maxime} \sur{Chamberland}}
\equalcont{These authors contributed equally to this work.}

\affil*[1]{\orgdiv{Department of Computer Science and Mathematics}, \orgname{Eindhoven University of Technology}, \orgaddress{\street{Groene Loper 5}, \city{Eindhoven}, \postcode{5612 AP}, \state{Noord-Brabant}, \country{The Netherlands}}}

\affil*[2]{\orgdiv{Department of Statistics}, \orgname{University of South Carolina}, \orgaddress{\city{Columbia}, \postcode{29225}, \state{South Carolina}, \country{United States of America}}}
\affil[3]{\orgdiv{Psychiatry Neuroimaging Laboratory}, \orgname{Brigham and Women’s Hospital, Harvard Medical School}, \orgaddress{\street{399 Revolution Drive}, \city{Boston}, \postcode{02215}, \state{MA}, \country{United States}}}


\abstract{Diffusion Magnetic Resonance Imaging (dMRI) is a powerful tool for probing brain microstructure, but clinical acquisitions are often limited by low out-of-plane resolution, resulting in degraded structural information and reduced utility for advanced analysis. We introduce SIINR (Structurally Informed Implicit Neural Representations), a general framework for super-resoltion of clinical dMRI datasets while quantifying uncertainty in the reconstructed outputs. SIINR utilizes a supervised 3D U-net as a prior and combines it with a self-supervised implicit neural representation (INR) that fuses the high-resolution prior and the original low-resolution data. The INR enables joint modeling across spatial and angular domains, enforces data consistency, and provides analytic approximate posterior distributions for downstream uncertainty quantification. We validate the framework on a diverse set of open-access dMRI datasets, demonstrating that SIINR outperforms standard interpolation methods in both quantitative error metrics and qualitative anatomical fidelity. Experiments on clinical cases, including subjects with multiple sclerosis and brain lesions, illustrate the framework’s ability to propagate intensity changes and flag uncertain regions in challenging scenarios. SIINR is flexible, modular, and can be adapted to different upsampling ratios and downstream tasks, providing a principled approach for enhancing clinical dMRI and supporting robust interpretation of derived neuroimaging metrics.}


\maketitle

\section{Introduction}\label{sec1}
Diffusion Magnetic Resonance Imaging (dMRI) is used to non-invasively probe the micro- and macrostructure of the brain \citep{basser1994mr}. High-quality acquisitions are used in research settings to fit increasingly complex microstructural models of the brain \citep{alexander2019imaging,  novikov2019quantifying, consagra2025deep}, and to create detailed visualizations of the white matter nerve fiber tracts, through a process called tractography \citep{malcolm2010filtered, jeurissen2019diffusion}. Obtaining these high-quality acquisitions, however, is a time-consuming process, with advanced dMRI protocols taking over an hour of in-scanner time \citep{van2013wu}. In clinical settings, these kinds of protocols are impractical. 
A common way to shorten the acquisition times is to decrease the out-of-plane resolution (i.e., the resolution in the axial direction). In-plane (i.e., in the transverse plane) resolution is kept reasonably high, which still allows the clinician to assess pathophysiological processes using dMRI.

The low out-of-plane resolution (i.e., thick slice acquisitions) causes an asymmetric loss of structural information. This negatively impacts quantification of dMRI metrics (e.g., fractional anisotropy or mean diffusivity) as well as the quality of the tractography results \citep{mcmaster2025sensitivity}, as they are naturally dependent on the structure, and in the case of severely anisotropic voxels might even make tractography impossible. Microstructural quantification can also be influenced, as larger voxels increase the partial voluming effects (i.e., effects caused by the presence of multiple different compartments with distinct diffusion profiles in a single voxel \cite{vos2011partial}). Essentially, the possibility of (scientific) analysis of the clinical dMRI acquisitions is limited because of the structural degradation, which, combined with the fact that clinical dMRI acquisitions are made relatively frequently, means that a wealth of information could be going to waste. 
A way to reintroduce the structure into the acquisitions would unlock new options for analyzing the images, and potentially lead to new scientific insights, especially for pathological cases that are often missing from the publicly available dMRI datasets.

Reintroducing structure in a low-resolution dMRI dataset is a super-resolution task. Many different strategies for single image super-resolution exist, ranging from standard interpolation (i.e., linear, cubic splines), to advanced deep-learning based methods \citep{karimiDiffusionMRIMachine2024a, luoDiffusionMRISuperresolution2022, chatterjeeShuffleUNetSuperResolution2021, kebiriThroughPlaneSuperResolutionAutoencoders2022, ordinolaSuperresolutionMappingAnisotropic2025, remediosECLAREEfficientCrossplanar2026}. While some \citep{alexander2017image} have attempted to use supervised learning methods to increase the spatial resolution on dMRI data, they have been limited to very small super resolution factors (2 or less) in the slice dimension. Within the domain of dMRI specifically, we are not aware of any method that deals effectively with excessively anisotropically degraded structure (super-resolution factor of 4). Super-resolution in dMRI is a complicated task, since its acquisitions consist of multiple correlated volumes. This usually prevents the full acquisition from being processed by the model at once during fitting because of GPU memory constraints, requiring either angular or spatial partitioning. This is problematic, since in both domains there are correlations that can be leveraged. Furthermore, supervised deep-learning are known to project training set biases onto unseen data, and they are usually inflexible when it comes to the shape of the input data. An additional concern with existing super-resolution methods, is that they do not provide a measure of uncertainty on the outputs they produce. Such a measure is necessary to provide confidence in the metrics that are calculated on the super-resolved datasets. These uncertainty estimations are unique to every subject, as they depend on the quality acquisition and similarity of it to the training set. Having such a measure allows the model to 'fail gracefully' as is often stated as a requirement for successful implementation into practice \citep{haldarStateArtMR2026}.

Previously, implicit neural representations (INRs) have been used for continuous functional representation and uncertainty quantification tasks in dMRI \citep{consagraNeuralOrientationDistribution2024a, hendriksNeuralSphericalHarmonics2023a, hendriksImplicitNeuralRepresentationCSD2025a, hendriksImplicitNeuralRepresentationsSM2025b}. INRs are self-supervised coordinate-based multi-layer perceptrons (MLPs) with a positional encoding technique, and are fit for each acquisition individually, i.e.,circumventing the training set bias. This results in a neural network that represents the full spatial and angular signal of a single subject implicitly in the weights and biases of the network. In the task of super-resolution, the INRs show good performance for small upsampling factors, but fail when the factor becomes too large. 

In this work, we introduce SIINR (Structurally Informed Implicit Neural Representations), a flexible framework designed to restore structural detail in clinical-resolution dMRI datasets. More importantly, we introduce it as a general framework that can be adapted to suit specific needs. Our proposed approach relies on a dual-component architecture: a supervised super-resolution model that serves as a soft prior, and an INR that grounds the outputs in the original acquisition while enabling uncertainty quantification. Together, these components reliably upsample clinical data, ensure fidelity to the raw scan, and provide explicit insight into the trustworthiness of the generated outputs.

\section{Methods}\label{methods}
In the following subsections we will conceptually introduce the problem \ref{methods:struct}, the parts of our framework \ref{method:frame}, and the implementation details \ref{method:impl}.
\subsection{Loss of structure}
\label{methods:struct}
In this section, we will phrase the loss of structure in a low-resolution sampling, compared to a high-resolution sampling, as an inverse problem. Given a volume $\Omega \in \mathbb{R}^3$ of a measurable quantity, we can take samples at different resolutions 
by dividing $\Omega$ into smaller subvolumes. The measurable quantity of interest in diffusion MRI is (directional)
movement of water molecules, which is expressed as signal (attenuation) and is a scalar value $y \in \mathbb{R_+}$ for a given measurement direction. 
Consider a low-resolution subvolume $\Omega^{lr}$, and consider a set of high-resolution volumes covering $\Omega^{lr}$, 
$\mathbf{\Omega^{hr}} = \{\Omega^{hr}_j | \sum^{N_j}_{j=1}\Omega^{hr}_j \supseteq \Omega^{lr}\}$, with $N_j$ being the number of subvolumes. 
The signal $y_{\Omega^{lr}}$ in $\Omega^{lr}$ can be seen as a weighted sum over signal $y_{\Omega^{hr}}$ in the smaller subvolumes
\begin{equation}
    y_{\Omega^{lr}} = \sum_{\mathbf{\Omega^{hr}}}\omega_jy_{\Omega^{hr}_j} 
    \label{eq:downsample}
\end{equation}
where  $\omega_j$ is a weighting determined by the slice profile of the low-resolution sample, and depends on the relative position of the high-resolution samples to the low-resolution sample.
We define a dMRI measurement as a set of non-overlapping equally sized subvolumes covering $\Omega$, $S = \{\Omega_1, \cdots, \Omega_N | \bigcup_{i=1}^N\Omega_i = \Omega\}$ where $N$ is the number of 
samples and $\Omega_i$ the subvolume.
We can take two samples covering $\Omega$ at different resolutions, the high-resolution set $S^{hr}$ with $N=N_{hr}$ and the low-resolution set $S^{lr}$ with $N=N_{lr}$, $N_{lr} < N_{hr}$. By (\ref{eq:downsample}), we can write every subvolume in the low-resolution
samples as a weighted sum of the high-resolution samples. Thus, we can obtain a
mapping $f: S^{hr} \rightarrow S^{lr}$. In the case of super resolution, we are
essentially interested in the inverse $f^{-1}$. This is an ill-posed problem, as many
different values of $y_{\Omega^{hr}_j}$ (even for known $\omega_j$) lead to the same $y_{\Omega^{lr}}$. In other words, structure in high-resolution cannot be recovered at a voxel-level without incorporating additional knowledge.
\subsection{Framework}
\label{method:frame}
\subsubsection{U-net}
By assuming there are identifiable patterns (such as location, and neighboring voxel intensities) in the low-resolution sampling that can be used to infer the underlying high-resolution structure, one could try to approximate $f^{-1}$. A (by now) straightforward way is trying to approximate $f^{-1}$ from known pairs of low- and high-resolution samplings by using supervised deep learning. In this work we make use of a standard 3D convolutional U-Net, which provides a flexible function approximator for imaging data. The trainable convolutional kernels integrate direct neighborhood information, while the multi-layer down- and upsampling architecture effectively increases the number of scales at which the model operates \citep{ronnebergerUNetConvolutionalNetworks2015}.
In dMRI, the signal in every subvolume depends on the location $\mathbf{v}$, which we define as the centerpoint of the subvolume, the measurement direction  $\mathbf{\hat{n}}$, and the diffusion 
weighting $\mathbf{b}$. By separating the dMRI acquisition into volumes of individual measurement directions, we revert to dealing with scalar values for each mapping, as across each volume $\mathbf{\hat{n}}$ and $\mathbf{b}$ are constant.
Thus, after training the U-net, we have $f^{-1} \approx \mathcal{U}_{\boldsymbol{\theta}_u} :S^{lr} \rightarrow S^{hr}$, 
where $\mathcal{U}_{\boldsymbol{\theta}_u}$ is the U-net parameterized by $\boldsymbol{\theta}_u$. Although this strategy decreases the information the U-net has for inferring the mapping, it has the benefit of making the learned mapping independent of the acquisition parameters $\mathbf{\hat{n}}$ and $\mathbf{b}$. 
It also enables using full volumes at once as training samples, without running into GPU memory constraints on consumer-grade hardware. Since this framework upsamples the volumes independently, we do not leverage the correlation between the volumes in a full acquisition. In the next paragraph we describe how we use INRs to mediate this issue, and provide further motivation for their use. 
\subsubsection{INR}
An INR is a multi-layer perceptron (MLP) that maps input coordinates to a desired output. They have been set-up in many ways, and have been used to represent a variety of signals. Recently, INRs have found their way into the domain of dMRI \citep{hendriksImplicitNeuralRepresentationCSD2025a, consagraNeuralOrientationDistribution2024a}. To avoid the spectral bias (i.e., the tendency of an MLP to learn low-frequency representations more easily), the INR framework transforms the coordinates into a high-dimensional frequency domain, either through explicit sinusoidal encodings before entering the model, or implicitly by using sinusoidal activation functions. In this work we will use an INR to learn a map from 3D coordinates into functions on the unit sphere representing the dMRI signal at a coordinate. 
$g : \mathbb{R}^3 \rightarrow \mathbb{S}^2 $.\\
The purpose of using an INR is threefold. First, by fitting all volumes of an acquisition, it finds a representation that takes into account correlations in the signal in the spherical as well as the spatial domain. This allows us to reintroduce correlations that are potentially lost by using a U-net. Second, the INR framework provides a natural approach to balancing the importance of the high-resolution output of the Unet with the actual low-resolution acquisition. Finally, the INR framework can be used to parameterize a Gaussian process, which provides an analytical function for an approximate posterior distribution on the signal, as shown by \cite{consagraNeuralOrientationDistribution2024a}. This posterior distribution can be used for uncertainty estimation both on the signal and on downstream metrics. Now follows an elaboration on how this posterior is obtained.\\
\subsubsection{Posterior distribution calculation}
\label{met:uq}
What follows is an adaptation of previous work by Consagra et al. \citep{consagraNeuralOrientationDistribution2024a}. To avoid repetition, we have only included parts of the derivation that are essential to understand the changes made to the original model. For further details on the derivation and set-up of the Gaussian process, please consult the original work, particularly Appendix A.\\
We model the diffusion signal as a function-valued random field $g(\mathbf{v}, \mathbf{\mathbf{\hat{n}}})$, such that for every spatial location $\mathbf{v} \in \Omega$ we can obtain a distribution over the spherical functions depending on the measurement direction $\mathbf{\hat{n}}$. We formulate these spherical functions as a truncated rank K symmetrical real-valued spherical harmonics series, as is common practice in dMRI. The signal is then given by
\begin{equation}
\label{eq:signal}
    g(\mathbf{v},\mathbf{\hat{n}}) = \sum_{k=0}^K \phi_k(\mathbf{\hat{n}})c_k(\mathbf{v}),
\end{equation}
where $c_k(\mathbf{v})$ is the coefficient for spherical harmonics basis function $k$ and $\phi_k(\mathbf{\hat{n}})$ is the spherical harmonics basis function $k$ evaluated in direction $\mathbf{\hat{n}}$. We will use an MLP to map spatial coordinates into coefficients vectors $\mathcal{M}_{\boldsymbol{\theta}_m}:\mathbf{v} \rightarrow \mathbf{c}$, where $\theta_m$ are the MLP parameters and $\mathbf{c}:= (c_1, ..., c_k)$ are the coefficients. Given equation \ref{eq:signal}, we can then describe our signal as
\begin{equation}
    g(\mathbf{v}, \mathbf{\hat{n}}) = \boldsymbol{\phi}(\mathbf{\hat{n}})\mathcal{M}_{\boldsymbol{\theta}_m}(\mathbf{v}),
\end{equation}
where $\boldsymbol{\phi} := (\phi_1, ..., \phi_k)$.

For a set of low-resolution measurements $\mathbf{V} \subset \Omega$ with coordinates $\mathbf{v}_1,..., \mathbf{v}_{M_c} \in \mathbf{V}$ (i.e., $M_c$ coordinates), we observe a signal in each of the $M_d$ directions $\{\mathbf{\hat{n}}_j|j\in1,...,M_d\}$, $\mathbf{y}_i = (y_{i,_1},...,y_{i,{M_d}})$. From Consagra et al. we see that the per-voxel likelihood for the low-resolution signal can be written as
\begin{equation}
\label{eq:pv_likelihood}
    \mathbf{y}_i|\mathbf{v}_i,\boldsymbol{\theta}_m,\sigma_e^2 \sim \mathcal{N}(\boldsymbol{\Phi}\mathcal{M}_{\boldsymbol{\theta}_m}(\mathbf{v_i}),\sigma_e^2\mathbf{I}),
\end{equation}
where $\sigma^2_e$ is the variance of the measurement error, $\boldsymbol{\Phi} \in \mathbb{R}^{M_d\times K}$ is the sample matrix constructed from basis functions and sampling directions, with $\boldsymbol{\Phi}_{jk} = \phi_k(\mathbf{\hat{n}}_j)$, and $\mathbf{I}$ is the identity matrix. In practice the value of $\sigma_e^2$ is estimated from the $b=0$ volumes if available, or set empirically (as in \cite{consagraNeuralOrientationDistribution2024a})..

Within the context of super-resolution, it is useful to interpret the low-resolution signal as the result of a downsampling operator applied to some higher resolution sampling. Let $\mathbf{v}_i$ be the center point of low-resolution voxel $i$, which occupies spatial volume $\Omega_i\subset\Omega$. We form the Monte-Carlo approximation of the downsampling operator 
$$
\frac{1}{|\Omega_i|}\int_{\Omega_i}\boldsymbol{\Phi}\mathcal{M}_{\boldsymbol{\theta}_m}(\mathbf{v})d\boldsymbol{v} \approx \frac{1}{M_{s}}\sum_{l=1}^{M_{s}}\boldsymbol{\Phi}\mathcal{M}_{\boldsymbol{\theta}_m}(\mathbf{v}_{il}); \quad \mathbf{v}_{il}\sim \text{Unif}(\Omega_i),
$$
with $M_s$ being the number of samples and $\text{Unif}(\Omega_i)$ denoting a uniformly distributed sample within volume $\Omega_i$.
Inserting this into (\ref{eq:pv_likelihood}) we obtain
\begin{equation}
    \mathbf{y}_i|\mathbf{v}_i,\boldsymbol{\theta}_m,\sigma_e^2 \sim \mathcal{N}(\frac{1}{M_s}\sum_{l=1}^{M_s}\boldsymbol{\Phi}\mathcal{M}_{\boldsymbol{\theta}_m}(\mathbf{v}_{il}),\sigma_e^2\mathbf{I}),
\end{equation}
where the variance is assumed to be $\sigma_e^2$ in each of the subsamples. Given the full dataset $\mathbf{Y} = [\mathbf{y}_1^\intercal, ...,\mathbf{y}_{M_c} ^\intercal] \in \mathbb{R}^{M_d\times M_c}$ and $\boldsymbol{\mathcal{M}}_{\boldsymbol{\theta}_m} = [\frac{1}{M_s}\sum_{l=1}^{M_s}\mathcal{M}_{\boldsymbol{\theta}_m}(\mathbf{v}_{1l}),...,\frac{1}{M_s}\sum_{l=1}^{M_s}\mathcal{M}_{\boldsymbol{\theta}_m}(\mathbf{v}_{M_cl})]^\intercal \in \mathbb{R}^{K\times M_c}$ the dataset likelihood then follows as 
\begin{equation}
    \mathbf{Y}|\mathbf{V}, \boldsymbol{\theta}_m, \sigma_e^2 \sim \mathcal{N}_{M_d\times M_c}(\boldsymbol{\Phi}\boldsymbol{\mathcal{M}}_{\boldsymbol{\theta}_m},\sigma_e^2\mathbf{I}_{M_d}, \mathbf{I}_{M_c}),
\end{equation}
where $\mathcal{N}_{M_d\times M_c}$ denotes the matrix normal distribution.\\
It has been shown \citep{consagraNeuralOrientationDistribution2024a} that by treating the INRs final layer random variables with a matrix normal prior distribution $\mathbf{W}\in \mathbb{R}^{r\times K}$, where $r$ is the output size of the INR up to the final layer, we can obtain an analytic vectorized distribution for the conditional posterior given by:
\begin{equation}
    \text{vec}(\mathbf{W})|\mathbf{V},\mathbf{Y},\boldsymbol{\theta}_w,\psi, \sigma_w^2, \sigma_e^2 \sim \mathcal{N}_{Kr}(\frac{1}{\sigma_e^2}\boldsymbol{\Lambda}_{\boldsymbol{\theta}_w}^{-1}[\boldsymbol{\mathcal{M}}^{\intercal}_{\boldsymbol{\theta}_w}\otimes\boldsymbol{\Phi}]^\intercal\text{vec}(\mathbf{Y}), \boldsymbol{\Lambda}_{\boldsymbol{\theta_w}}^{-1})
\end{equation}
where $\boldsymbol{\theta}_w$ are the INR parameters up to $\mathbf{W}$, $\psi$ is used to compute the spectral density of the angular prior (again, see \cite{consagraNeuralOrientationDistribution2024a} for details) on $\mathbb{S}^2$, $\sigma_w^2$ is the prior variance on $\mathbf{W}$, $\boldsymbol{\mathcal{M}}_{\boldsymbol{\theta}_w} = [\frac{1}{M_s}\sum_{l=1}^{M_s}\mathcal{M}_{\boldsymbol{\theta}_w}(\mathbf{v}_{1l}),...,\frac{1}{M_s}\sum_{l=1}^{M_s}\mathcal{M}_{\boldsymbol{\theta}_w}(\mathbf{v}_{M_cl})]^\intercal \in \mathbb{R}^{r\times M_c}$ with $\mathcal{M}_{\boldsymbol{\theta}_w}$ being the INR up to $\mathbf{W}$, $\otimes$ denotes the Kronecker product, and 
\begin{equation}
    \boldsymbol{\Lambda}_{\boldsymbol{\theta_w}} = \frac{1}{\sigma_e^2}(\frac{\sigma_e^2}{\sigma_w^2}\mathbf{I}_r \otimes \mathbf{R}_\psi + \boldsymbol{\mathcal{M}}_{\boldsymbol{\theta}_w}\boldsymbol{\mathcal{M}}^{\intercal}_{\boldsymbol{\theta}_w} \otimes \boldsymbol{\Phi}^\intercal\boldsymbol{\Phi}).
\end{equation}
with $\mathbf{R}_\psi$ being a diagonal matrix of spectral densities for different spherical harmonics orders. By linearity, it follows that 
\begin{equation}
\label{eq:post}
\begin{aligned}
    \boldsymbol{c}(\boldsymbol{v}) | \mathbf{V},\mathbf{Y},\boldsymbol{\theta}_w,\psi, \sigma_w^2, \sigma_e^2  \sim \mathcal{N}\Big(&\frac{1}{\sigma_e^2}\left[\boldsymbol{\mathcal{M}}_{\theta_{w}}^{\intercal}(\boldsymbol{v}) \otimes\boldsymbol{I}_{K}\right]^{\intercal}\boldsymbol{\Lambda}_{\boldsymbol{\theta}_w}^{-1}[\boldsymbol{\mathcal{M}}^{\intercal}_{\boldsymbol{\theta}_w}\otimes\boldsymbol{\Phi}]^\intercal\text{vec}(\mathbf{Y}), \\
    &\left[\boldsymbol{\mathcal{M}}_{\theta_{w}}^{\intercal}(\boldsymbol{v})\otimes \boldsymbol{I}_{K}\right]\boldsymbol{\Lambda}_{\theta_{w}}^{-1}\left[\boldsymbol{\mathcal{M}}_{\theta_{w}}^{\intercal}(\boldsymbol{v})\otimes \boldsymbol{I}_{K}\right]^{\intercal}\Big).
\end{aligned}
\end{equation}
Having obtained the distribution over $\boldsymbol{c}(\boldsymbol{v})$ we can obtain distributions over downstream metrics by sampling from $\boldsymbol{c}(\boldsymbol{v})$, and proceeding with downstream calculations using each sample. 

\subsection{Implementation details}
\label{method:impl}
In this section we describe the collection of the dataset for supervised learning, as well as specify the implementation details and motivate the choices made during implementation.
\subsubsection{Supervised learning dataset}\label{methods:dataset}
In order to cover a broad distribution of possible dMRI inputs while training the U-net, we combine a number of existing open-access datasets.
From the Human Connectome Project Young Adult (HCP-YA) dataset we include 100 subjects scanned on a Connectom 3T scanner, and 50 subjects scanned on the 7T scanner, with a base resolution of 1.25 mm and 1.05 mm isotropic, respectively, and $b=1000 s/mm^2$ \citep{vanessenWUMinnHumanConnectome2013c}. From the HCP Early Psychosis (HCP-EP) dataset we include 100 subjects with $b=500 s/mm^2$, 
and 50 subjects with $b=1500 s/mm^2$, both with a base resolution of 1.5 mm isotropic \citep{jacobsIntroductionHumanConnectome2025}. From the HCP Aging Adult Brain Connectome (HCP-AABC) we include 100 subjects aged over 60, with a base resolution of 1.5 mm isotropic and $b=1500 s/mm^2$ \citep{bookheimerLifespanHumanConnectome2019}. Finally, from the Diffusion MRI Harmonisation Challenge we included 14 subjects with state-of-the-art acquisitions on the Prisma (CDMRI-P) and Connectom (CDMRI-C) scanners \citep{taxCrossscannerCrossprotocolDiffusion2019a}. This results in a dataset with 428 subjects for a total of 37,820 diffusion weighted or $b=0 s/mm^2$ volumes. We apply a 70-20-10 train-validation-test split, based on subject to prevent data leakage. An overview of the full dataset is shown in Table \ref{table:data}.

\begin{table}[htbp]
\resizebox{\textwidth}{!}{\begin{tabular}{l|rrrrrrr}
Name        & HCP-YA & HCP-YA & HCP-EP& HCP-EP& HCP-AABC & CDMRI-P & CDMRI-C    \\
\hline
\# Subjects & 100      & 50  & 100 & 50       & 100   &    14 & 14               \\
Age avg     & $\sim$25 & $\sim$25 & $\sim$23 & $\sim$23  & \textgreater60 & - & -\\
Resolution  & 1.25     & 1.05 & 1.5 & 1.5 & 1.5 & 1.5 & 1.2                    \\
B-value     & 1000     & 1000 & 500 & 1500 & 1000 & 1200 & 1200                  \\
TR          & 5520     & 7000 &  3230 & 3230 & 8800 & 7200  &  7200  \\
TE          & 89.5 & 71.2  & 89.22 & 89.22 & 57 &  80       &    68    \\
\# dir      & 108 & 79 &  12 & 187 & 107 & 65 & 65                       \\
Field & 3T & 7T & 3T & 3T & 3T & 3T & 3T
\end{tabular}}
\caption{Description of the datasets used to create the training dataset for the supervised network.}
\label{table:data}
\end{table}

\subsubsection{U-net implementation}
The U-net in this framework adheres to a basic 3D U-net structure, with three downsampling layers each halving the resolution,
a latent space convolution followed by three upsampling layers each doubling the resolution, with
skipped connections between layers at the equal levels. All layers use $3\times3\times3$ kernels with
32 output channels in the first layer, doubling every deeper layer.

\begin{figure}
    \centering
    \includegraphics[width=\textwidth]{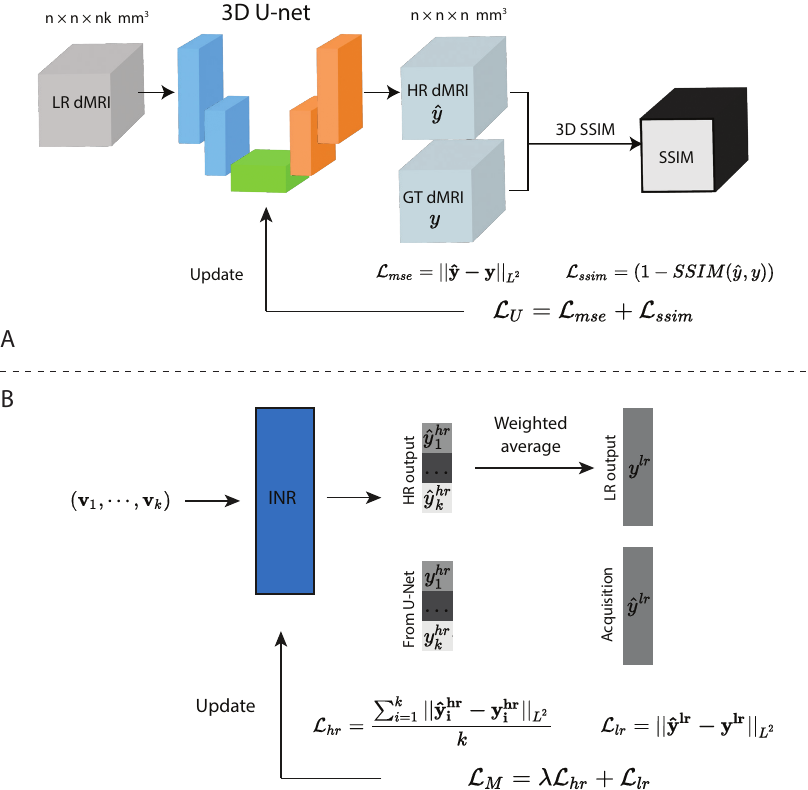}
    \caption{The SIINR framework architecture. Part A shows the supervised U-net, which is trained using a structural similarity (SSIM) and an MSE loss. Part B shows the INR fit which is performed for an individual subject, which is updated using a combined MSE loss on the high-resolution U-net output, and the low-resolution acquisition.}
    \label{fig:architecture}
\end{figure}

The loss function is made up of two components. A conventional voxel-wise mean-squared error (MSE) loss $\mathcal{L}_{mse}$, 
and a structural similarity index measure (SSIM) loss $\mathcal{L}_{ssim}$
\begin{align}
        \text{SSIM}(\hat{y}, y) &= \frac{(2\mu_{\hat{y}}\mu_y+c_1)(2\sigma_{\hat{y}y} + c_2)}{(\mu^2_{\hat{y}} + \mu^2_y + c_1)(\sigma_{\hat{y}}^2 + \sigma_y^2 + c_2)}\\
        \mathcal{L}_{\text{ssim}} &= 1-\text{SSIM}(\hat{y},y)
\end{align}
where $\mathbf{\hat{y}}$ is the 3D model output, $\mathbf{y}$ is the 3D target, $\mu$ and $\sigma$ denote the mean and (co-)variance over a gaussian window of size $w\times w \times w$ voxels, and $c_1$ and $c_2$ are variables to stabilize the division. We choose the default values of $c_1 = 0.01^2$ and $c_2 = 0.03^2$, and a window size of $k=11$. SSIM provides a way to quantify structural similarity between two images, in contrast to the point-wise nature of the MSE. Using $\mathcal{L}_{\text{ssim}}$
can lead to more accurate reconstructions of the target image \citep{ghodratiMRImageReconstruction2019}. Since structure is our main interest in learning, using $\mathcal{L}_{\text{ssim}}$ is a natural choice. The complete loss function $\mathcal{L}$ then is
\begin{equation}
    \mathcal{L}_U = \mathcal{L}_{\text{mse}} + \mathcal{L}_{\text{ssim}}.
\end{equation}
For stability reasons, the first 500 volumes are fit using only $\mathcal{L}_{\text{mse}}$, afterwards $\mathcal{L}_{\text{ssim}}$ is added. Since $\mathcal{L}_{\text{ssim}}$ generally is larger than $\mathcal{L}_{\text{mse}}$, it becomes the domination factor in the loss function.
An Adam optimizer with a learning rate of 1e-4 is used, decaying with a factor 0.8 every epoch. Gradients are accumulated over 5 volumes, before being used to update the model parameters. The model is fit for 8 epochs, after which the model with the best performance on the validation set is chosen. The architecture can be seen in Figure \ref{fig:architecture}A.

\subsubsection{INR implementation}
The INR consists of a four layer MLP $\mathcal{M}_{\boldsymbol{\theta}_m}$ with a hidden dimension size of 1024 and ReLU activation functions on all layers but the output layer. The output $\mathbf{c}$ has a size which depends on the maximum order of the SH representation of the signal $l_{max}$, and is given by $\frac{1}{2}(l_{max} + 1 )(l_{max} + 2)$. The number of inputs is determined by the hyperparameter $n_{p}$, which determines the size of the matrix $\mathbf{A}$ used in the Fourier feature encoding
\begin{equation}
    \gamma(\mathbf{v}) = [\cos(2\pi\mathbf{A}\mathbf{v}), \sin(2\pi\mathbf{A}\mathbf{v})]
    \label{eq_enc}
\end{equation}
where $\gamma(.)$ is the Fourier feature encoding, and $\mathbf{A}$ is a size $n_p \times 3$ matrix with values sampled from $\mathcal{N}(0, \sigma^2)$. The variance $\sigma^2$ is a hyperparameter that can be used to adapt the spatial smoothness of the output. Before encoding the inputs normalized to lie within $[-1,1]^3$.  The inputs $\gamma(\mathbf{v}) \in [-1, 1]^{n_p\times 2}$ are therefore a vector of size $n_p\times2$.\\
The parameters $\boldsymbol{\theta}_m$ of the MLP are updated by computing the MSE between the measured signal and the output. Since we fit the MLP with two, potentially conflicting targets, the loss function consists of two parts. The first part is the MSE loss between the output signal, and the high-resolution U-net output $\mathcal{L}_{hr}$ (the prior). The second part is a MSE loss between the weighted sum of the high-resolution outputs and the measured low-resolution data $\mathcal{L}_{lr}$ (see Equation~\ref{eq:downsample})). The complete loss function is then given by
\begin{equation}
    \mathcal{L}_{M}=\lambda\mathcal{L}_{hr} + \mathcal{L}_{lr},
\end{equation}
where $\lambda$ is a weighting parameter to balance the importance of the high- and low-resolution loss. In practice, to keep error and output magnitudes similar between the high- and low-resolution volumes, we make use of the weighted average rather than the weighted sum. The INR is fit to each dataset in 250 epochs, with an Adam optimizer with a 1e-4 learning rate. The hyperparemeter $\sigma^2$ is set to 3.5, $n_p$ to 5000, and $l_{max}$ to 8 across experiments. The architecture can be seen in Figure \ref{fig:architecture}B.

\section{Evaluation}
First we will define the downstream tasks that will be used to evaluate the performance of our framework. Then, we will discuss the metrics that are being used for quantitative analysis. Next, we describe the comparison methods. Finally, we define the experiments by describing which data is used for which tasks.
\subsection{Diffusion signal reconstruction quality}
\label{ex:diff_rec}
For each subject the high-resolution dMRI dataset is reconstructed, using the SIINR framework, from a low-resolution version that is either the original acquired resolution, or created by downsampling 4$\times$ in z-direction. The high-resolution SIINR outputs are compared quantitatively on reconstruction quality with the original high-resolution dataset, if available, by computing the normalized MSE (see Section~\ref{metrics}). Qualitative analysis consists of visual inspection of signal and error maps.

\subsection{DTI metrics}
The apparent diffusion coefficient (ADC) and fractional anisotropy (FA) are calculated from diffusion tensors fit on high-resolution reconstructions from \ref{ex:diff_rec}. The diffusion tensors are fit to the datasets using DiPY version 1.11.0 \citep{garyfallidisDipyLibraryAnalysis2014}. The quality is quantitatively evaluated using RMSE compared to the ADC and FA calculated on the original high-resolution data, if available. Qualitative analysis consists of visual inspection of ADC and FA and their error maps.

\subsection{Fiber orientation distributions}
\label{ex:fod}
The fiber orientation distributions (FODs) are obtained through constrained spherical deconvolution (CSD) using MRtrix3 \citep{tournierRobustDeterminationFibre2007b, tournierMRtrix3FastFlexible2019a}. The quality of the FODs is quantitatively evaluated by computing the apparent fiber density (AFD), which is defined as the average over the FOD or, equivalently, the first coefficient in the SH representation (zero order coefficient). Qualitative analysis consists of visual inspection of the FODs, compared to the other methods.

\subsection{Fiber tracking}
\label{ex:fibertrack}
The FODs obtained in section \ref{ex:fod} are used for fiber tracking. Bundle reconstructions were performed for one representative subject across three reconstruction methods (i.e., ground truth, cubic interpolation, and implicit neural representation). Tract-specific masks (i.e., corpus callosum, corticospinal tracts) and corresponding inclusion regions defining tract endpoints were obtained using TractSeg \citep{wasserthalTractSegFastAccurate2018a} on the high-resolution DWI volumes reconstructed by the different methods. Streamline tractography was then generated in FiberNavigator following the protocol of Chamberland et al. \citep{chamberland2014real}, using one seed per voxel, a step size of 1 mm, an angular threshold of 40°, and an FOD amplitude threshold of 0.1. The same tracking parameters were applied across all reconstruction conditions to ensure a fair comparison of the resulting tractograms.

\subsection{Uncertainty quantification}
Using the methods described in Section~\ref{met:uq}, we compute a posterior distribution over the SH coefficients of the signal using equation \ref{eq:post}. From these distributions we obtain 250 realizations by Monte-Carlo sampling. These realizations are used in downstream tasks, providing a posterior distribution over the downstream output. We quantify the uncertainty per output by calculating the coefficient of variation (CVA, see Section \ref{metrics}). The quality of the uncertainty quantification is assessed by visual inspection of the CVA maps, and comparing them to error maps of downstream tasks.

\subsection{Comparisons}
As additional comparisons, the low-resolution dataset is upsampled to high-resolution using linear and cubic interpolation. The outputs are processed in the same way as the SIINR output described above. Both are commonly used methods for increasing the resolution of low-resolution scans before performing downstream tasks.

\subsection{Metrics}
\label{metrics}
For the signal, we are interested in evaluating the quality of all volumes of the combined dMRI data set. Since the signal intensities vary across different subjects and b-values, we normalize the error by the 95-th percentile of the original acquisition. The MSE ($L_2$-norm) per volume is computed and averaged over the subject
\begin{align}
    \text{MSE}_i &= ||\mathbf{\hat{y}_i-\mathbf{y_i}}{}||_{L^2}\\
    \text{MSE}_{\text{DWI}}&=\frac{1}{M_d}\sum_{i=1}^{M_d}\frac{\text{MSE}_i}{P^{95}_i},
\end{align}
where $\mathbf{\hat{y}_i}$ is the estimated value for volume position $i$, $\mathbf{y_i}$ is the measured value for volume $i$, $M_d$ is the number of volumes, and $P^{95}_i$ is the 95th percentile of volume $i$. 
The ADC, FA and AFD maps are compared to the maps calculated on the acquired signal using the RMSE (i.e., $\sqrt{\text{MSE}}$), which are then averaged over the whole volume. The RMSE on the DWI and FA volumes are calculated within a brain mask. The RMSE AFD is calculated within three white matter regions, as the results of CSD are only valid inside of the white matter (see Section~\ref{ex:healthy}). The white matter ROIs (i.e., corpus callosum, cingulum, and corticospinal tract), were chosen as they primarily run in x, y, and z directions. Binary maps for these regions were created from the high-resolution acquisition using TractSeg. We also show fractional error maps created by a voxel-wise division of the error by the mean value.  These maps allow us to see how large the error is compared to the size of the ground truth. 

For uncertainty quantification, we make use of the voxel-wise coefficient of variation, which is a unitless metric that quantifies the width of the distribution relative to its mean 
\begin{equation}
    \text{CVA}_i = \frac{\sigma_i}{\mu_i},
\end{equation}
where $\sigma_i$ is the variance of the distribution and $\mu_i$ the mean in voxel $i$.

\section{Experiments}
\subsection{Experiment 1: Framework performance on test set}
\label{ex:healthy}
To validate our framework's performance on samples within the distribution of the training data, we make use of the unseen test set of the dataset described under Section~\ref{methods:dataset}. For each subject, the mean of the metrics is calculated over the regions of interest (i.e., full brain or tracts). The average across all subjects is then computed and shown. Scalar maps are shown for one of the worst, and one of the best performing outputs. The FODs and fiber tracts are qualitatively compared for selected subjects. Uncertainty quantification is performed for a subject with higher MSE to get an indication of its usefulness in detecting wrong reconstructions.

\subsection{Experiment 2: Intensity and morphological changes }
To evaluate the performance of the algorithm on "out-of-distribution" data that have abnormal intensity and/or morphological changes not present in the training data, ADC and FA are calculated for two clinical subjects with different pathologies. Both acquisitions are acquired at a low out-of-plane resolution. Uncertainty quantification is also performed and visualized.

\subsection{Experiment 3: U-net vs. SIINR}
Here we provide qualitative comparison between our "prior" the U-net outputs and those obtained subsequently from the SIINR outputs, to assess the benefits of using an INR to represent the full dataset, outside of enabling uncertainty quantification. On a representative subject, we compute the ADC and FA maps using the U-net outputs directly, and the SIINR outputs. The results are qualitatively analyzed, where differences between the U-net and SIINR outputs are highlighted.

\section{Results}\label{results}
\subsection{Experiment 1: Test set performance}
\label{exp1:test}
\subsubsection{Quantitative results}
The quantitative comparisons for the DWI signal, FA, and AFD are shown in Table \ref{table:quant}. We see a lower error for all metrics and datasets for the SIINR framework, compared to linear and cubic interpolation.

\begin{table}[htbp]
\caption{Quantitative results for the test dataset, averages and standard deviations are shown for different source datasets and all datasets combined. (R)MSE = (root) mean squared error, DWI = diffusion weighted image, ADC is apparent diffusion coefficient, FA = fractional anisotropy, AFD = apparent fiber density.}
\hspace{10px}
\begin{tabular}{r|rrrr}
\multicolumn{1}{l|}{Dataset} & DWI MSE & ADC RMSE & FA RMSE & AFD RMSE\\
\hline
\multicolumn{1}{l|}{HCP-YA 3T} & & & \\
\it{Linear} & 0.0219 ± 0.0014& 0.00018 ± 2.7e-05& 0.0755 ± 0.0024& 0.0254 ± 0.0117 \\
\it{Cubic} & 0.0199 ± 0.0013& 0.00018 ± 2.7e-05& 0.0676 ± 0.0022& 0.0228 ± 0.0102 \\
\it{SIINR} & 0.0129 ± 0.0018& 0.00013 ± 2.6e-05& 0.0485 ± 0.002& 0.0141 ± 0.0029 \\
\multicolumn{1}{l|}{HCP-YA 7T} & & & \\
\it{Linear} & 0.0312 ± 0.0038& 0.00019 ± 2.1e-05& 0.0984 ± 0.0066& 0.0219 ± 0.009 \\
\it{Cubic} & 0.0279 ± 0.0034& 0.00018 ± 2.1e-05& 0.0881 ± 0.006& 0.0191 ± 0.0073 \\
\it{SIINR} & 0.0158 ± 0.0015& 0.00015 ± 1.4e-05& 0.0734 ± 0.007& 0.0091 ± 0.0029 \\
\multicolumn{1}{l|}{HCP-EP b500} & & & \\
\it{Linear} & 0.0128 ± 0.0024& 0.0002 ± 5.9e-05& 0.1172 ± 0.0093& 0.0214 ± 0.0102 \\
\it{Cubic} & 0.0115 ± 0.0021& 0.00019 ± 5.7e-05& 0.1056 ± 0.0088& 0.0198 ± 0.0092 \\
\it{SIINR} & 0.0082 ± 0.0013& 0.00018 ± 6.3e-05& 0.0817 ± 0.0074& 0.015 ± 0.0065 \\
\multicolumn{1}{l|}{HCP-EP b1500} & & & \\
\it{Linear} & 0.0216 ± 0.0027& 9e-05 ± 5e-06& 0.0694 ± 0.0027& 0.0221 ± 0.0126 \\
\it{Cubic} & 0.0199 ± 0.0025& 9e-05 ± 5e-06& 0.0615 ± 0.0027& 0.0199 ± 0.0109 \\
\it{SIINR} & 0.0179 ± 0.0021& 7e-05 ± 5e-06& 0.0444 ± 0.0016& 0.0135 ± 0.0061 \\
\multicolumn{1}{l|}{HCP-AABC} & & & \\
\it{Linear} & 0.0231 ± 0.0023& 0.00021 ± 6.4e-05& 0.0758 ± 0.0037& 0.023 ± 0.0065 \\
\it{Cubic} & 0.0208 ± 0.0021& 0.0002 ± 6.8e-05& 0.0685 ± 0.0036& 0.0211 ± 0.0061 \\
\it{SIINR} & 0.0123 ± 0.0014& 0.00016 ± 6.5e-05& 0.0498 ± 0.0043& 0.0133 ± 0.004 \\
\multicolumn{1}{l|}{CDMRI-P} & & & \\
\it{Linear} & 0.0269 ± 0.0019& 0.00023 ± 1.1e-05& 0.09 ± 0.0047& 0.0321 ± 0.0036 \\
\it{Cubic} & 0.0247 ± 0.0017& 0.00022 ± 1.2e-05& 0.0805 ± 0.004& 0.0302 ± 0.0032 \\
\it{SIINR} & 0.0172 ± 0.0017& 0.00018 ± 1.1e-05& 0.062 ± 0.0016& 0.0147 ± 0.0032 \\
\multicolumn{1}{l|}{CDMRI-C} & & & \\
\it{Linear} & 0.0279 ± 0.0015& 0.00018 ± 9e-06& 0.0839 ± 0.003& 0.0139 ± 0.0064 \\
\it{Cubic} & 0.0253 ± 0.0015& 0.00017 ± 9e-06& 0.0741 ± 0.003& 0.0129 ± 0.0065 \\
\it{SIINR} & 0.0169 ± 0.0016& 0.00013 ± 5e-06& 0.0584 ± 0.0024& 0.0105 ± 0.0036 \\
\multicolumn{1}{l|}{Combined} & & & \\
\it{Linear} & 0.0223 ± 0.0062& 0.00018 ± 5.4e-05& 0.0881 ± 0.0178& 0.0229 ± 0.0102 \\
\it{Cubic} & 0.0202 ± 0.0056& 0.00018 ± 5.4e-05& 0.079 ± 0.0162& 0.0209 ± 0.0091 \\
\it{SIINR} & 0.0134 ± 0.0036& 0.00015 ± 5.4e-05& 0.0601 ± 0.0147& 0.0132 ± 0.0049 \\
\end{tabular}

\label{table:quant}
\end{table}

\subsubsection{Qualitative results of DWI, ADC, and FA}
\label{exp1:dwiadc}
\begin{figure}[htbp]
    \centering
    \includegraphics[width=\textwidth]{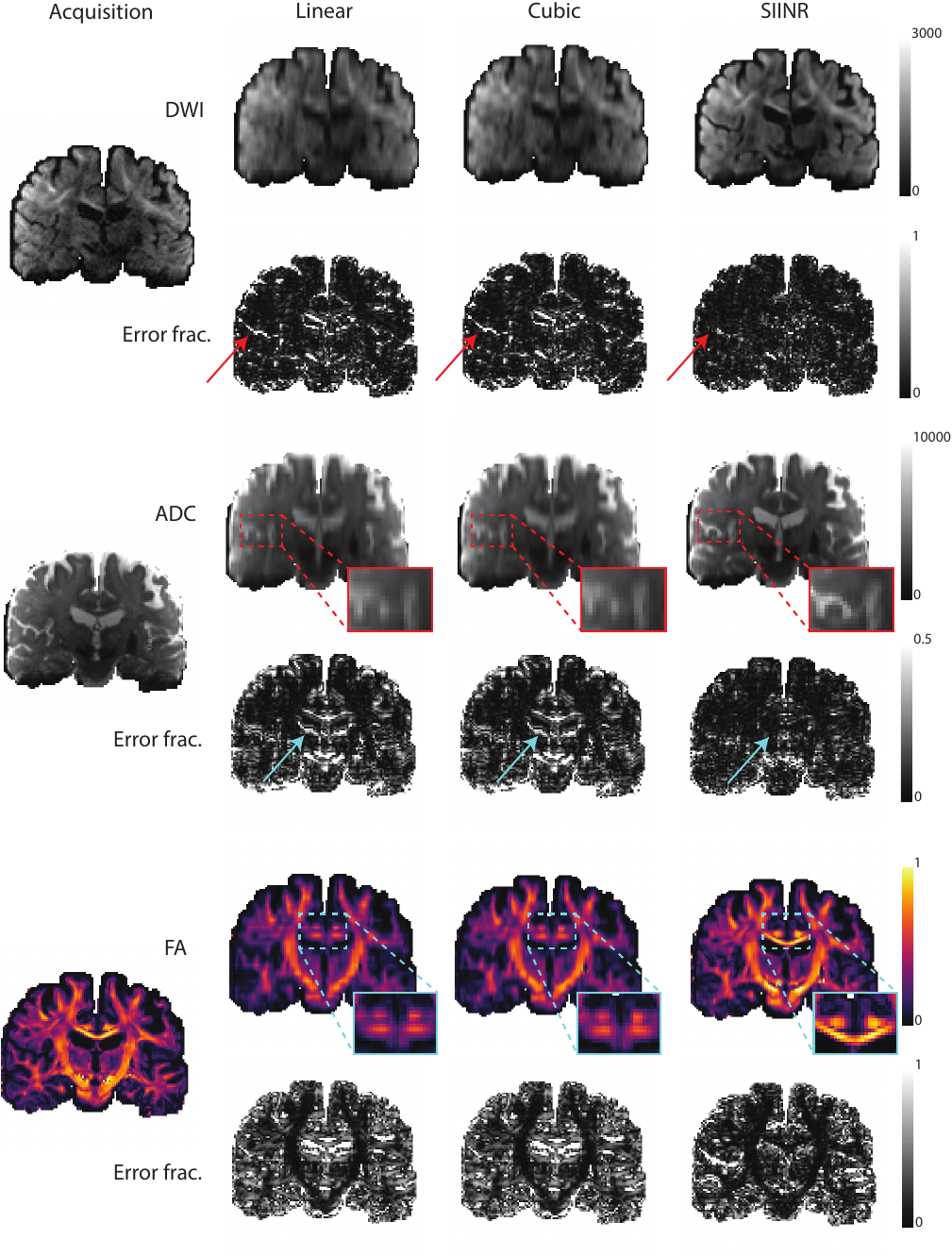}
    \caption{Coronal slices of a DWI volume, the ADC, and FA, for a representative subject from the test set. We note a clear improvement in sharpness in the SIINR output compared to linear and cubic interpolation (red and blue inlays). The fractional error maps show a decrease in error, particularly at tissue and corticospinal-fluid interfaces (red and blue arrows). Interpolation artifacts (i.e., broken/split corpus callosum) are visible in both linear and cubic interpolation (blue inlays).}
    \label{fig:testset}
\end{figure}

We randomly selected a subjected from the AABC dataset (with a 1.5 mm isotropic original resolution) to perform this analysis. For a slice of the upsampled DWI volumes, the ADC volume, and the FA, we computed the error fraction compared to the high-resolution acquisition. The results are shown in Figure \ref{fig:testset}. As can be seen, major anatomical discrepancy is seen if naive linear or cubic interpolation is used with the corpus callosum appearing interrupted. However, the proposed SIINR output is very similar to the ground truth data, with no anatomical discrepancy.
\FloatBarrier

\subsubsection{Qualitative results of FODs and fiber tracking}
\begin{figure}
    \centering
    \includegraphics[width=\textwidth]{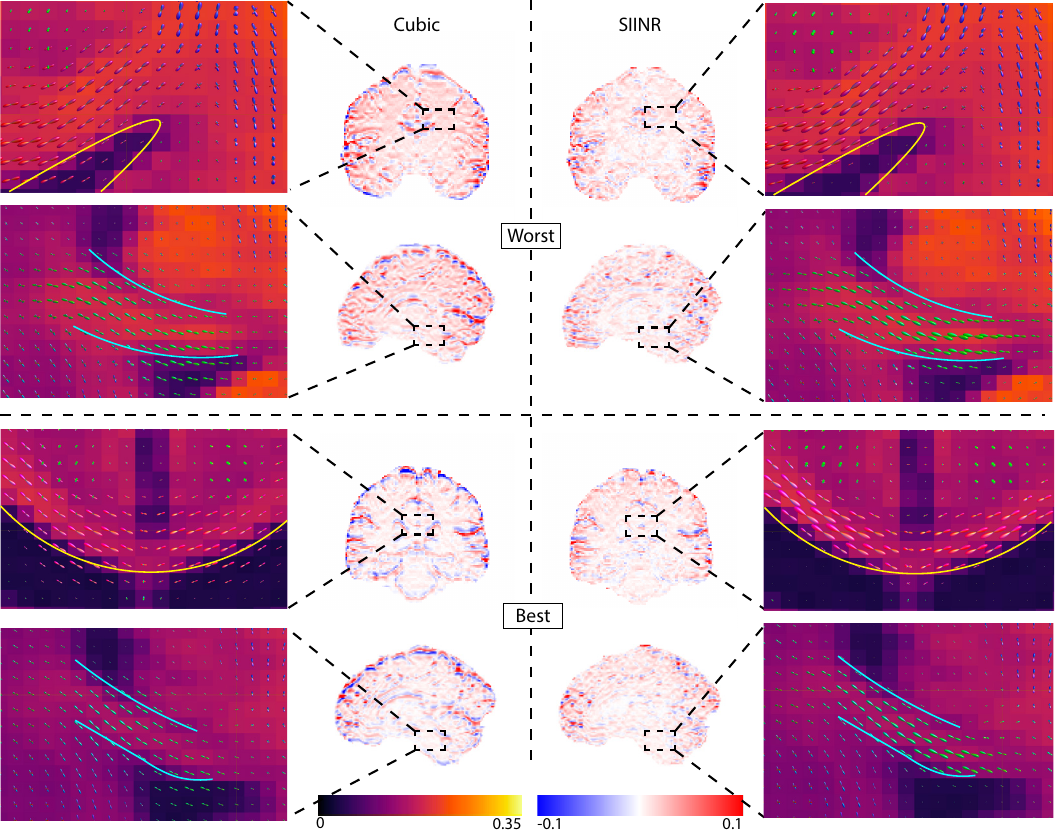}
    \caption{AFD error maps and FODs in regions of interest for two subjects. In the error maps (center columns), red indicates overestimation of the AFD, blue indicates underestimation. We note an overall higher error for cubic interpolation. Around the corpus callosum we can clearly see partial voluming effects, where FODs extend across the anatomical boundaries (yellow lines). These effects are also present around the cerebellar peduncle (blue lines). The FODs are shown against a background of AFD calculated on the high-resolution acquired data.}
    \label{fig:fod}
\end{figure}

\begin{figure}
    \centering
    \includegraphics[width=\textwidth]{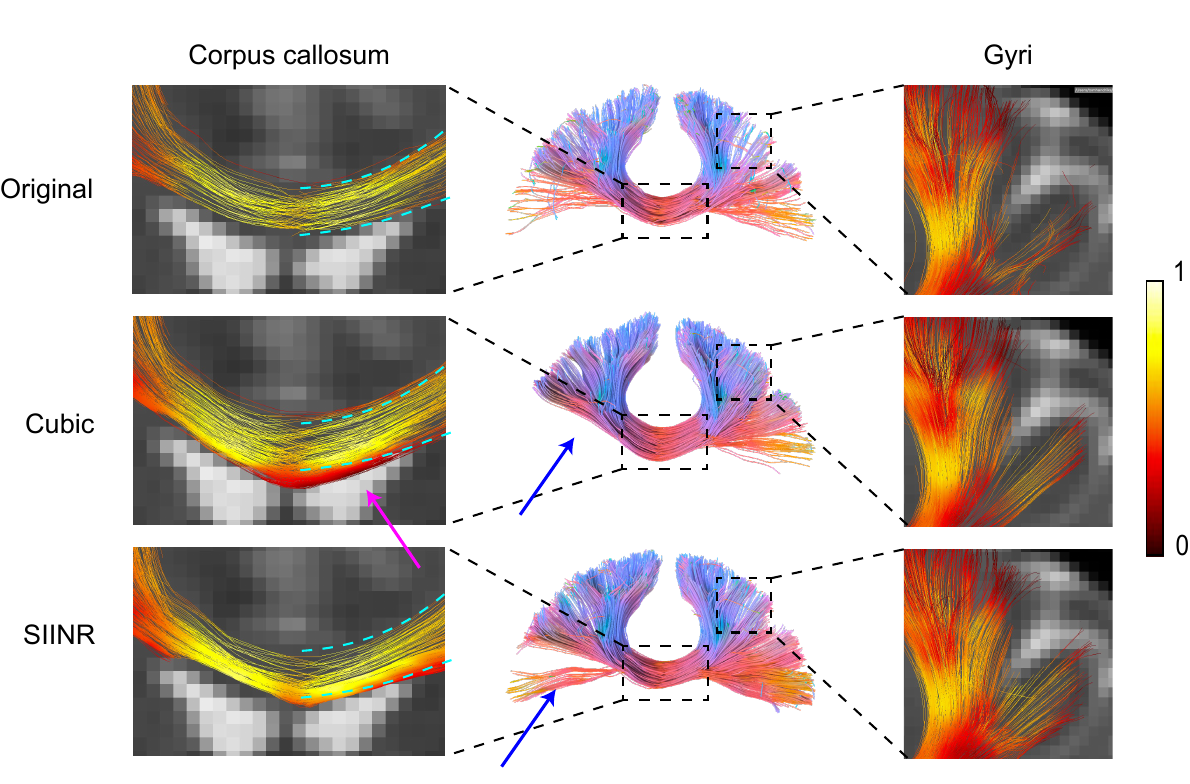}
    \caption{Results of the tractography of the corpus callosum (CC). The 3D visualizations (center column) provides an overview of the entire bundle, color-coded by direction. The inlay images show a 2D representation, where only the streamlines within the slice are shown colored by FA calculated on the high-resolution acquisition. We note an overshoot of anatomical boundaries (blue dashed lines) in the CC region of the tractogram produced with the cubic interpolation data. In the gyri cubic interpolation produces unnaturally straight lines, while SIINR more naturally fills the gyrus.}
    \label{fig:tracking}
\end{figure}

For two selected subjects from the HCP-EP b1500 dataset (i.e., one with the best performance of SIINR compared to cubic interpolation, and one with the worst), difference maps for AFD are shown alongside FOD visualizations for both cubic interpolation and the SIINR framework in Figure \ref{fig:fod}. The results of fiber tracking the corpus callosum are also shown for the worst performing subject in Figure \ref{fig:tracking}. Both figures illustrate how cubic interpolation suffers from partial voluming effects, or loss of structure, which results in the FODs and streamlines crossing anatomical boundaries. The FODs and streamlines produced by SIINR are better confined to within these boundaries.

\subsubsection{Uncertainty quantification}
\begin{figure}
    \centering
    \includegraphics[width=\textwidth]{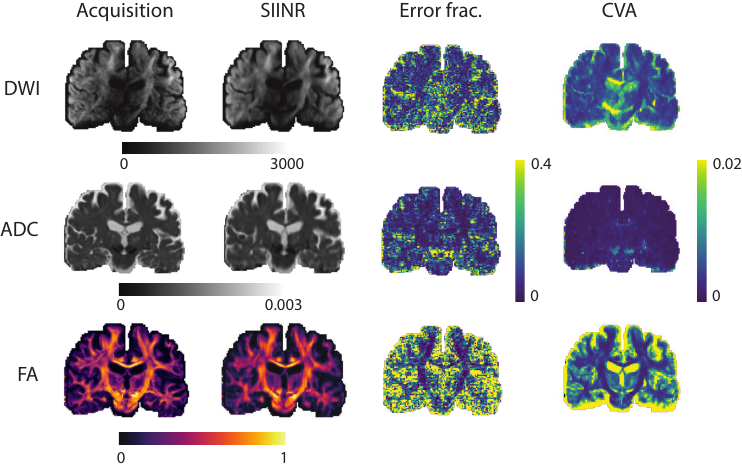}
    \caption{Coronal slices of diffusion signal (DWI), apparent diffusion coefficient (ADC), and fractional anisotropy (FA) shown alongside the error fraction and coefficient of variation (CVA). There are clear similarities the localizations of higher error fraction and higher CVA, although the error fraction tends to be higher in a area that is slightly larger than the areas of the higher CVA.}
    \label{fig:uq}
\end{figure}

Uncertainty quantification for a representative subject, i.e., the same as in Section \ref{exp1:dwiadc} for three different tasks: DWI signal reconstruction, ADC calculation and FA calculation are shown in Figure \ref{fig:uq}.

\subsection{Experiment 2: Clinical data}
\subsubsection{Multiple sclerosis}
\begin{figure}
    \centering
    \includegraphics[width=\textwidth]{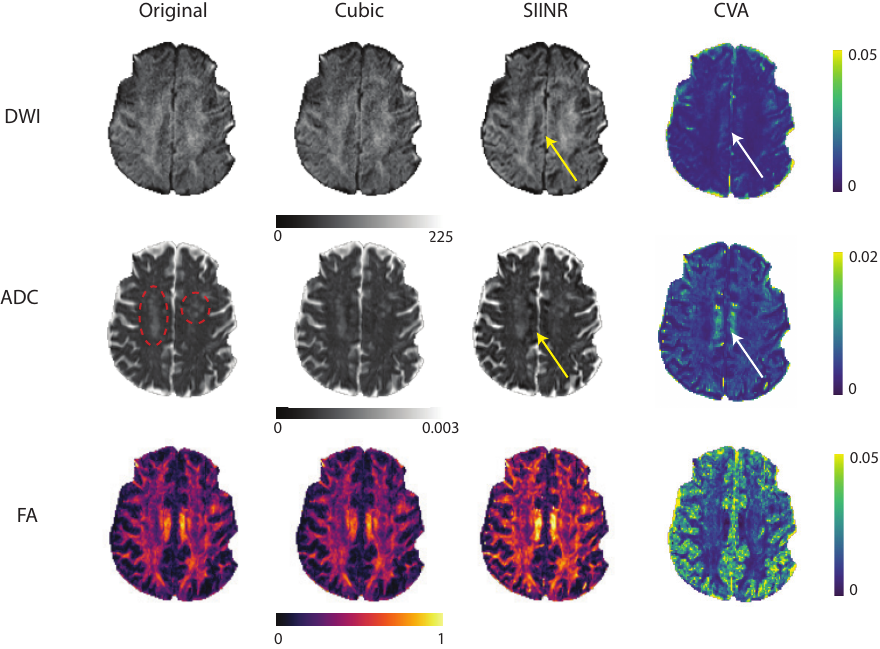}
    \caption{The DWI and parameter maps from a clinical dataset of a subject with multiple sclerosis (MS). The slice in cubic and SIINR interpolations is the high-resolution slice closest to the center of the low-resolution slice. The MS lesions can be clearly seen on the apparent diffusion coefficient (ADC) map as hyperintensities in the white matter (red ellipses). Immediately medial to these lesions we see hypointensities in the DWI and the ADC on the SIINR (yellow arrows), which are less pronounced than in the original. These are indicated as high uncertainty (white arrows). The FA generally seems higher for the SIINR reconstruction than the original, which is expected given the smaller voxels. Uncertainty in FA is located mostly in the gray matter areas, which is expected due to the low anisotropy in these areas.}
    \label{fig:ms}
\end{figure}

Our first out-of-distribution testing is on a subject with multiple sclerosis (MS), which has a distinct MRI characteristic of periventricular white matter lesions. In $T_2$ weighted images, these show up as hyperintense areas. The original acquisition has 1.375x1.375x6 $mm^3$ resolution, with 24 $b=1000s/mm^2$ and 4 $b=0s/mm^2$ volumes. They are upsampled to 1.375x1.375x1.5 mm resolution using the SIINR framework and cubic interpolation. The resulting high-resolution datasets are then used for computing downstream metrics and uncertainty quantification. The results are shown in Figure \ref{fig:ms}.

\subsubsection{Lesion}
\begin{figure}
    \centering
    \includegraphics[width=\textwidth]{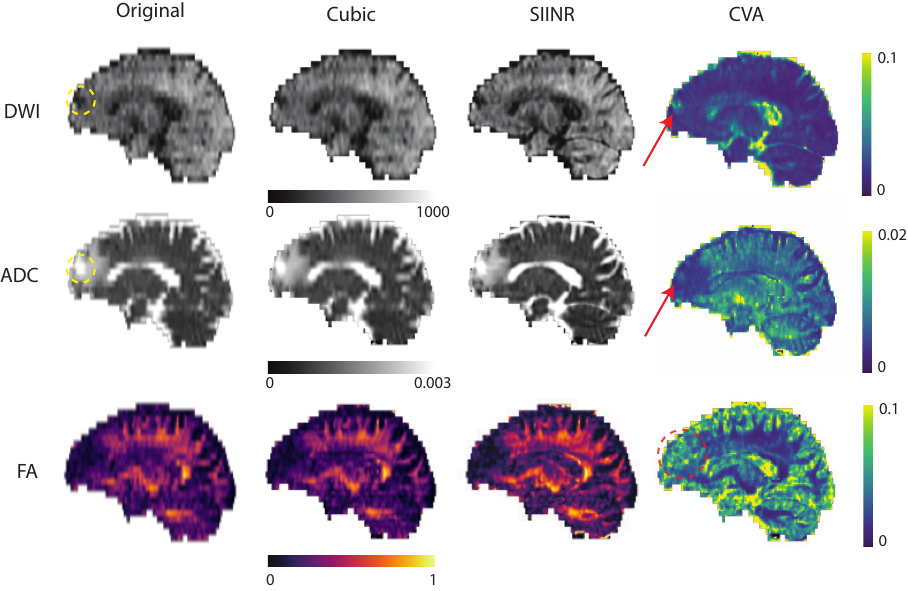}
    \caption{The DWI and parameter maps from clinical dataset of a subject with a frontal lobe lesion. The lesion core (yellow circle) is attenuated in the DWI slice and is bright on the ADC map. Surrounding the lesion, there is an area of edema which has high ADC and low FA. The lesion core shows up as an area of higher uncertainty in the DWI and ADC map. The entire lesion region, as well as the gray matter, shows higher uncertainty in the FA map.}
    \label{fig:tumor}
\end{figure}

This subject has a lesion/edema in the frontal lobe, which shows up on MRI as an abnormality in the structure. The original acquisition has 1.5x1.5x6 mm resolution, with 25 $b=1000s/mm^2$ and 1 $b=0s/mm^2$ volumes. This data is upsampled to 1.5 mm isotropic resolution using the SIINR framework and cubic interpolation. The resulting high resolution datasets are then used for computing downstream metrics. The results are shown in Figure \ref{fig:tumor}.
As can be seen, the proposed framework can handle unseen data well and preserves the anatomy of the lesion, and the signal intensity if the surrounding tissue. The uncertainty quantification shows that the model is confident in the ADC output in the tumor area, but is uncertain about the FA estimates.

\subsection{Experiment 3: Comparing U-net and SIINR}
\begin{figure}
    \centering
    \includegraphics[width=\textwidth]{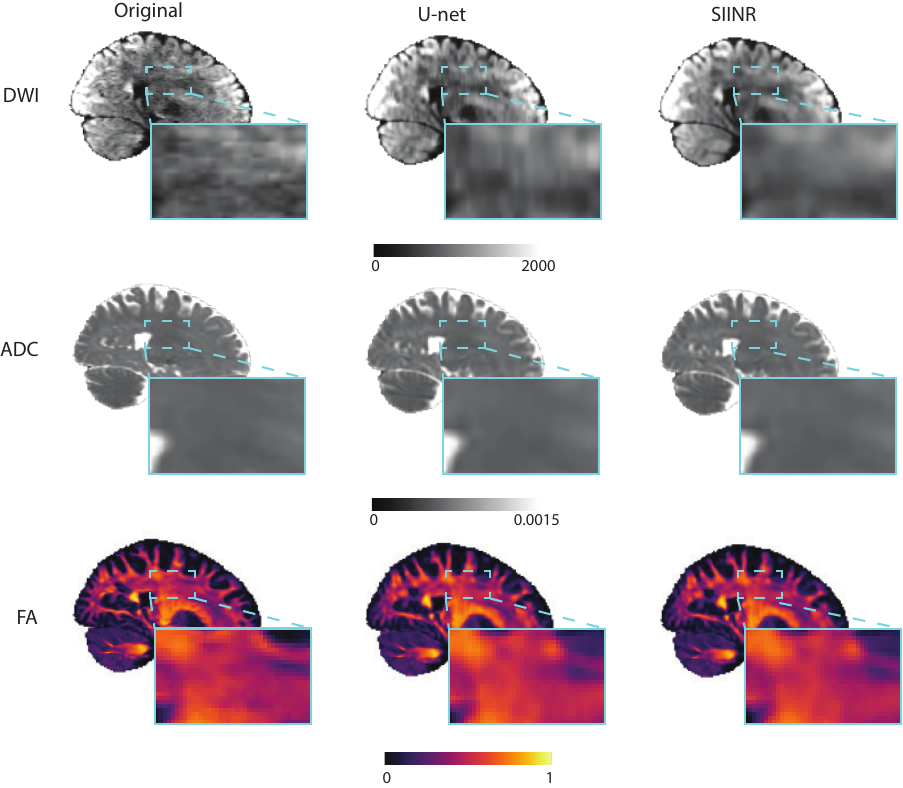}
    \caption{Comparison between U-net and SIINR outputs. In the individual DWI volumes there are block artifacts present in the U-net output, which do not appear in the SIINR output. After using all DWI volumes to compute ADC or FA, however, these artifacts are no longer present.}
    \label{fig:unet_ir}
\end{figure}

A representative subject was selected from the HCP-EP dataset. The DWI and parameter maps fit directly to the U-net output and fit on the output of SIINR can be seen in Figure \ref{fig:unet_ir}. The artifacts in the DWI reconstruction are significantly reduced by SIINR, while in the ADC and FA the differences are not immediately obvious.

\section{Discussion}\label{discussion}
\subsection{Framework Advantages and Real-world Applicability}
In this work, we present SIINR as a framework that enables users to faithfully improve the spatial resolution of their clinical datasets, so that all downstream tasks can be run using existing software pipelines. Additionally, it uniquely allows uncertainty quantification over the outputs and the derived metrics. This provides insight to the user on the reliability of their results, and allows the SIINR model to 'fail gracefully', as is necessary when applying it in real-world situations. The main goal of this work was not to obtain the highest performing model for every specific task, but rather to show the potential of the framework to perform super-resolution for dMRI data at high out-of-plane super-resolution factors.

The results on the test datasets from Section~\ref{exp1:test} show that SIINR consistently outperforms standard super-resolution methods, with qualitative assessments further revealing enhanced anatomical structure in the reconstructions. The uncertainty quantification shows uncertainty in areas where the reconstruction error is also larger. Prior work \citep{consagraNeuralOrientationDistribution2024a} using scan-rescan datasets further supports our approach.

While clinical datasets lack ground truth, they do show how the framework could be used in a real-world scenario. Intensity changes are properly propagated through the framework, and generally result in plausible looking results. For example, SIINR highlights uncertainty in regions where the supervised model struggles to reconstruct (such as near MS lesions in ADC maps). In the lesion case, the lesion core is marked by high uncertainty, while surrounding edema primarily impacts FA uncertainty.

We also compared the results of the U-net to the full framework. For individual DWI volumes there are pronounced differences, with the U-net showing clear block artifacts where SIINR does not. This improvement is likely due to 1) the INR’s ability to leverage spatial correlations and 2) its use of a truncated SH series, which inherently limits high-frequency changes across DWI volumes. The block artifacts are not visible in the downstream metrics, most likely because these are aggregations of all DWI volumes into a diffusion tensor which also does not allow for high frequency changes between DWI measurement directions. For other downstream tasks, or with fewer measurement directions the artifacts might become relevant. If the U-net output and the actual acquisition severely disagree, e.g., in the case of a hallucination by the U-net, the INR course-corrects and potentially highlights this in the uncertainty quantification, and in the disparity between both outputs. We have not encountered large hallucinations in this work by the U-net, so this remains to be investigated.

\subsection{Limitations of the Training Dataset and Degradation Approach}\label{discussion:dataset}
Supervised machine learning benefits from a training dataset that contains as much of the variation that is present in the target distribution. In the field of dMRI the variation is due to many factors. Scanner-related factors (brand, model, field strength, etc.), acquisition-related factors (B-tensor, number of acquisitions, echo-time, read-out technique, etc.), data-processing related factors (pre- and post-processing steps), subject-related factors (age, sex, presence of pathology, etc.), and possibly others. While SIINR’s design (i.e., U-net for super-resolution of each volume individually) offers flexibility to accommodate diverse acquisition schemes, it is infeasible to comprehensively cover all possible combinations of these factors in a single training dataset.

Our approach to simulating low-resolution data also introduces limitations. To align with conventions, we have used trilinear interpolation to degrade the scans to a lower resolution. This implies that signal intensities remain in the same range. While this facilitates model comparison and training, it does not fully mimic the physical process of acquiring data at lower resolution. The summation described in Section~\ref{methods:struct} more closely resembles the true underlying physics, but still is not identical to acquiring a volume at a lower resolution. 

\subsection{Interpretation of quantitative metrics}
In the absence of a true, noiseless ground truth, the quantitative metrics become difficult to interpret. Both the U-net and INR components of SIINR possess denoising capabilities, which may produce representations that more accurately reflect the underlying biological structures; however, this denoising effect can increase the error metrics when compared to the original, noisy acquisitions. The quantitative analysis of FODs is challenging in the absence of a true 'ground truth' fiber population, as they are very sensitive to the noise present in the original acquisition. Nonetheless, the AFD can give an indication of partial voluming effects, while the results of fiber tracking give a more qualitative insight into their usability. It is, furthermore, important to study the parameter maps of the outputs, to get insight into the distribution of the errors across the volume. Given the impracticality of reporting qualitative results for all slices, subjects, and metrics, we focused on a subset of subjects. Where possible, we have selected examples that do not represent the best-case performance, so as to avoid overstating the capabilities of our framework.

\subsection{Flexibility and Potential Improvements in the Supervised Super-resolution Model}
A central strength of the SIINR framework lies in its adaptability: the supervised model responsible for estimating $f^{-1}$ can be replaced with any super-resolution architecture suited to the user’s objectives. For specific reconstruction tasks, there may be alternative models that are better optimized than the standard U-net employed in this work. For example, tailoring a U-net to the precise upsampling ratio or anatomical region of interest could yield performance gains over a more generalized network. Regardless of the chosen supervised model, the INR component remains valuable for enforcing data consistency with the actual acquisition and for providing robust uncertainty quantification. In this work, we chose a relatively straightforward U-net implementation, which has been used successfully for many different tasks in the past including super-resolution. However, integrating more state-of-the-art models could further improve the results of the primary upsampling, and therefore the results of the SIINR framework overall. Ultimately, users are free to substitute their preferred or task-specific models within the SIINR pipeline.

\subsection{INR Hyperparameters and Task-specific Optimization}
The INRs used to model the dMRI datasets in this work all had identical hyperparameter settings. Earlier work has shown that, while being robust, these influence the results depending on the properties of the dataset (i.e., SNR, resolution, etc.) \cite{hendriksImplicitNeuralRepresentationCSD2025a, hendriksImplicitNeuralRepresentationsSM2025b, hendriksNeuralSphericalHarmonics2023a}. This suggests that there is considerable potential for fine-tuning INR hyperparameters to optimize performance for particular downstream tasks or specific data properties.

\section{Conclusion}
We present SIINR, a general and flexible framework for performing super-resolution of thick-slice clinical dMRI data, consisting of a supervised learning model that upsamples the low-resolution data, combined with an INR that uses data consistency to ensure robust data recovery while allowing for uncertainty quantification. While our current implementation serves as a proof-of-concept, the framework is readily adaptable to diverse clinical and research scenarios, supporting a range of super-resolution ratios and downstream analysis tasks.

\section*{Acknowledgements}
HCP data were provided [in part] by the Human Connectome Project, WU-Minn Consortium (Principal Investigators: David Van Essen and Kamil Ugurbil; 1U54MH091657) funded by the 16 NIH Institutes and Centers that support the NIH Blueprint for Neuroscience Research; and by the McDonnell Center for Systems Neuroscience at Washington University.\\
Research reported in this publication was supported by the National Institute On Aging of the National Institutes of Health under Award Number U01AG052564 and by funds provided by the McDonnell Center for Systems Neuroscience at Washington University in St. Louis. The HCP-Aging [and AABC] data used in this report came from ConnectomeDB powered by BALSA: https://balsa.wustl.edu/project?project=AABC.\\
Research using Human Connectome Project for Early Psychosis (HCP-EP) data reported in this publication was supported by the National Institute of Mental Health of the National Institutes of Health under Award Number U01MH109977. The HCP-EP 1.1 Release data used in this report came from DOI: 10.15154/1522899.
The CDMRI data were acquired at the UK National Facility for In Vivo MR Imaging of Human Tissue Microstructure located in CUBRIC funded by the EPSRC (grant EP/M029778/1), and The Wolfson Foundation. Acquisition and processing of the data was supported by a Rubicon grant from the NWO (680-50-1527), a Wellcome Trust Investigator Award (096646/Z/11/Z), and a Wellcome Trust Strategic Award (104943/Z/14/Z). This database was initiated by the 2017 and 2018 MICCAI Computational Diffusion MRI committees (Chantal Tax, Francesco Grussu, Enrico Kaden, Lipeng Ning, Jelle Veraart, Elisenda Bonet-Carne, and Farshid Sepehrband) and CUBRIC, Cardiff University (Chantal Tax, Derek Jones, Umesh Rudrapatna, John Evans, Greg Parker, Slawomir Kusmia, Cyril Charron, and David Linden).

\section *{Data and Software statement}
All data used in this work is publicly available, with exception of the two clinical acquisitions. For the clinical subjects, informed consent was obtained from a subject as per the requirements of MGBs IRB.
All code used in this work is publicly available through GitHub: \url{https://github.com/tomhend/SIINR-public}.

\section*{Declarations}
The authors declare to have no conflicts of interests.

\section*{Funding}
This work is made possible by the ISMRM Research Exchange Grant awarded to TH in 2025.
MC acknowledges funding from the Dutch Research Council (NWO, OCENW.M.22.352).

\section*{Contributions}
Conceptualization: TH, WC, AV, YR, MC; Data curation: TH, YR; Formal analysis: TH, WC; Funding acquisition: TH, YR, AV, MC; Investigation: TH, WC, YR, MC; Methodology: TH, WC, AV, YR, MC; Resources: AV, YR, MC; Software: TH, WC; Supervision: WC, AV, YR, MC; Validation: TH, WC; Visualization: TH, WC, AV, YR, MC; Writing – original draft: TH, WC; Writing – review and editing: TH, WC, AV, YR, MC.

\begin{appendices}

\end{appendices}

\bibliography{sn-bibliography}

@article{alexander2019imaging,
  title={Imaging brain microstructure with diffusion MRI: practicality and applications},
  author={Alexander, Daniel C and Dyrby, Tim B and Nilsson, Markus and Zhang, Hui},
  journal={NMR in Biomedicine},
  volume={32},
  number={4},
  pages={e3841},
  year={2019},
  publisher={Wiley Online Library}
}

@article{malcolm2010filtered,
  title={Filtered multitensor tractography},
  author={Malcolm, James G and Shenton, Martha E and Rathi, Yogesh},
  journal={IEEE Transactions on Medical Imaging},
  volume={29},
  number={9},
  pages={1664--1675},
  year={2010},
  publisher={IEEE}
}

@article{chamberland2014real,
  title={Real-time multi-peak tractography for instantaneous connectivity display},
  author={Chamberland, Maxime and Whittingstall, Kevin and Fortin, David and Mathieu, David and Descoteaux, Maxime},
  journal={Frontiers in neuroinformatics},
  volume={8},
  pages={59},
  year={2014},
  publisher={Frontiers Media SA}
}

@article{alexander2017image,
  title={Image quality transfer and applications in diffusion MRI},
  author={Alexander, Daniel C and Zikic, Darko and Ghosh, Aurobrata and Tanno, Ryutaro and Wottschel, Viktor and Zhang, Jiaying and Kaden, Enrico and Dyrby, Tim B and Sotiropoulos, Stamatios N and Zhang, Hui and others},
  journal={NeuroImage},
  volume={152},
  pages={283--298},
  year={2017},
  publisher={Elsevier}
}

@article{consagra2025deep,
  title={A deep learning approach to multi-fiber parameter estimation and uncertainty quantification in diffusion MRI},
  author={Consagra, William and Ning, Lipeng and Rathi, Yogesh},
  journal={Medical Image Analysis},
  volume={102},
  pages={103537},
  year={2025},
  publisher={Elsevier}
}

@article{vos2011partial,
  title={Partial volume effect as a hidden covariate in DTI analyses},
  author={Vos, Sjoerd B and Jones, Derek K and Viergever, Max A and Leemans, Alexander},
  journal={Neuroimage},
  volume={55},
  number={4},
  pages={1566--1576},
  year={2011},
  publisher={Elsevier}
}

@article{mcmaster2025sensitivity,
  title={Sensitivity of quantitative diffusion MRI tractography and microstructure to anisotropic spatial sampling},
  author={McMaster, Elyssa M and Newlin, Nancy R and Cho, Chloe and Rudravaram, Gaurav and Saunders, Adam M and Krishnan, Aravind R and Remedios, Lucas W and Kim, Michael E and Xu, Hanliang and Schilling, Kurt G and others},
  journal={Magnetic Resonance Imaging},
  pages={110539},
  year={2025},
  publisher={Elsevier}
}

@article{novikov2019quantifying,
  title={Quantifying brain microstructure with diffusion MRI: Theory and parameter estimation},
  author={Novikov, Dmitry S and Fieremans, Els and Jespersen, Sune N and Kiselev, Valerij G},
  journal={NMR in Biomedicine},
  volume={32},
  number={4},
  pages={e3998},
  year={2019},
  publisher={Wiley Online Library}
}

@article{jeurissen2019diffusion,
  title={Diffusion MRI fiber tractography of the brain},
  author={Jeurissen, Ben and Descoteaux, Maxime and Mori, Susumu and Leemans, Alexander},
  journal={NMR in Biomedicine},
  volume={32},
  number={4},
  pages={e3785},
  year={2019},
  publisher={Wiley Online Library}
}

@article{basser1994mr,
  title={MR diffusion tensor spectroscopy and imaging},
  author={Basser, Peter J and Mattiello, James and LeBihan, Denis},
  journal={Biophysical journal},
  volume={66},
  number={1},
  pages={259--267},
  year={1994},
  publisher={Elsevier}
}

@article{van2013wu,
  title={The WU-Minn human connectome project: an overview},
  author={Van Essen, David C and Smith, Stephen M and Barch, Deanna M and Behrens, Timothy EJ and Yacoub, Essa and Ugurbil, Kamil and Wu-Minn HCP Consortium and others},
  journal={Neuroimage},
  volume={80},
  pages={62--79},
  year={2013},
  publisher={Elsevier}
}

@article{chatterjeeShuffleUNetSuperResolution2021,
  title = {{{ShuffleUNet}}: {{Super}} Resolution of Diffusion-Weighted {{MRIs}} Using Deep Learning},
  shorttitle = {{{ShuffleUNet}}},
  author = {Chatterjee, Soumick and Sciarra, Alessandro and Dunnwald, Max and Mushunuri, Raghava Vinaykanth and Podishetti, Ranadheer and Rao, Rajatha Nagaraja and Gopinath, Geetha Doddapaneni and {Oeltze-Jafra}, Steffen and Speck, Oliver and Nurnberger, Andreas},
  year = 2021,
  month = aug,
  journal = {2021 29th European Signal Processing Conference (EUSIPCO)},
  pages = {940--944},
  publisher = {IEEE},
  address = {Dublin, Ireland},
  doi = {10.23919/EUSIPCO54536.2021.9615963},
  urldate = {2025-10-27},
  isbn = {9789082797060}
}

@article{karimiDiffusionMRIMachine2024a,
  title = {Diffusion {{MRI}} with Machine Learning},
  author = {Karimi, Davood and Warfield, Simon K.},
  year = 2024,
  month = nov,
  journal = {Imaging Neuroscience},
  volume = {2},
  pages = {imag-2-00353},
  issn = {2837-6056},
  doi = {10.1162/imag_a_00353},
  urldate = {2026-06-08}
}

@article{kebiriThroughPlaneSuperResolutionAutoencoders2022,
  title = {Through-{{Plane Super-Resolution With Autoencoders}} in {{Diffusion Magnetic Resonance Imaging}} of the {{Developing Human Brain}}},
  author = {Kebiri, Hamza and {Canales-Rodr{\'i}guez}, Erick J. and Lajous, H{\'e}l{\`e}ne and {de Dumast}, Priscille and Girard, Gabriel and {Alem{\'a}n-G{\'o}mez}, Yasser and Koob, M{\'e}riam and Jakab, Andr{\'a}s and Bach Cuadra, Meritxell},
  year = 2022,
  month = may,
  journal = {Frontiers in Neurology},
  volume = {13},
  publisher = {Frontiers},
  issn = {1664-2295},
  doi = {10.3389/fneur.2022.827816},
  urldate = {2025-10-27},
  langid = {english}
}

@article{luoDiffusionMRISuperresolution2022,
  title = {Diffusion {{MRI}} Super-Resolution Reconstruction via Sub-Pixel Convolution Generative Adversarial Network},
  author = {Luo, Suyang and Zhou, Jiliu and Yang, Zhipeng and Wei, Hong and Fu, Ying},
  year = 2022,
  month = may,
  journal = {Magnetic Resonance Imaging},
  volume = {88},
  pages = {101--107},
  issn = {0730-725X},
  doi = {10.1016/j.mri.2022.02.001},
  urldate = {2025-10-24}
}

@article{ordinolaSuperresolutionMappingAnisotropic2025,
  title = {Super-Resolution Mapping of Anisotropic Tissue Structure with Diffusion {{MRI}} and Deep Learning},
  author = {Ordinola, Alfredo and Abramian, David and Herberthson, Magnus and Eklund, Anders and {\"O}zarslan, Evren},
  year = 2025,
  month = feb,
  journal = {Scientific Reports},
  volume = {15},
  number = {1},
  pages = {6580},
  publisher = {Nature Publishing Group},
  issn = {2045-2322},
  doi = {10.1038/s41598-025-90972-7},
  urldate = {2025-10-24},
  copyright = {2025 The Author(s)},
  langid = {english}
}

@article{remediosECLAREEfficientCrossplanar2026,
  title = {{{ECLARE}}: Efficient Cross-Planar Learning for Anisotropic Resolution Enhancement},
  shorttitle = {{{ECLARE}}},
  author = {Remedios, Samuel W. and Wei, Shuwen and Han, Shuo and Zhang, Jinwei and Carass, Aaron and Schilling, Kurt G. and Pham, Dzung L. and Prince, Jerry L. and Dewey, Blake E.},
  year = 2026,
  month = mar,
  journal = {Journal of Medical Imaging},
  volume = {13},
  number = {2},
  pages = {024001},
  publisher = {SPIE},
  issn = {2329-4302},
  doi = {10.1117/1.JMI.13.2.024001},
  urldate = {2026-06-08},
  langid = {english}
}

@article{consagraNeuralOrientationDistribution2024a,
  title = {Neural Orientation Distribution Fields for Estimation and Uncertainty Quantification in Diffusion {{MRI}}},
  author = {Consagra, William and Ning, Lipeng and Rathi, Yogesh},
  year = 2024,
  month = apr,
  journal = {Medical Image Analysis},
  volume = {93},
  pages = {103105},
  issn = {1361-8415},
  doi = {10.1016/j.media.2024.103105},
  urldate = {2025-10-24}
}

@article{hendriksImplicitNeuralRepresentationCSD2025a,
  title = {Implicit Neural Representation of Multi-Shell Constrained Spherical Deconvolution for Continuous Modeling of Diffusion {{MRI}}},
  author = {Hendriks, Tom and Vilanova, Anna and Chamberland, Maxime},
  year = {2025a},
  month = mar,
  journal = {Imaging Neuroscience},
  volume = {3},
  pages = {imag\_a\_00501},
  issn = {2837-6056},
  doi = {10.1162/imag_a_00501},
  urldate = {2025-10-24}
}

@article{hendriksImplicitNeuralRepresentationsSM2025b,
  title = {Implicit Neural Representations for Accurate Estimation of the {{Standard Model}} of White Matter},
  author = {Hendriks, Tom and Arends, Gerrit and Versteeg, Edwin and Vilanova, Anna and Chamberland, Maxime and Tax, Chantal M. W.},
  year = {2025b},
  month = dec,
  journal = {Communications Biology},
  volume = {9},
  number = {1},
  pages = {120},
  publisher = {Nature Publishing Group},
  issn = {2399-3642},
  doi = {10.1038/s42003-025-09399-5},
  urldate = {2026-06-08},
  copyright = {2025 The Author(s)},
  langid = {english}
}

@inproceedings{hendriksNeuralSphericalHarmonics2023a,
  title = {Neural {{Spherical Harmonics}} for~{{Structurally Coherent Continuous Representation}} of~{{Diffusion MRI Signal}}},
  booktitle = {Computational Diffusion MRI},
  author = {Hendriks, Tom and Vilanova, Anna and Chamberland, Maxime},
  editor = {Karaman, Muge and Mito, Remika and Powell, Elizabeth and Rheault, Francois and Winzeck, Stefan},
  year = 2023,
  pages = {1--12},
  publisher = {Springer Nature Switzerland},
  address = {Cham},
  doi = {10.1007/978-3-031-47292-3_1},
  isbn = {978-3-031-47292-3},
  langid = {english}
}

@article{bookheimerLifespanHumanConnectome2019,
  title = {The {{Lifespan Human Connectome Project}} in {{Aging}}: {{An}} Overview},
  shorttitle = {The {{Lifespan Human Connectome Project}} in {{Aging}}},
  author = {Bookheimer, Susan Y. and Salat, David H. and Terpstra, Melissa and Ances, Beau M. and Barch, Deanna M. and Buckner, Randy L. and Burgess, Gregory C. and Curtiss, Sandra W. and {Diaz-Santos}, Mirella and Elam, Jennifer Stine and Fischl, Bruce and Greve, Douglas N. and Hagy, Hannah A. and Harms, Michael P. and Hatch, Olivia M. and Hedden, Trey and Hodge, Cynthia and Japardi, Kevin C. and Kuhn, Taylor P. and Ly, Timothy K. and Smith, Stephen M. and Somerville, Leah H. and U{\u g}urbil, K{\^a}mil and {van der Kouwe}, Andre and Van Essen, David and Woods, Roger P. and Yacoub, Essa},
  year = 2019,
  month = jan,
  journal = {NeuroImage},
  volume = {185},
  pages = {335--348},
  issn = {1053-8119},
  doi = {10.1016/j.neuroimage.2018.10.009},
  urldate = {2026-06-08}
}

@article{jacobsIntroductionHumanConnectome2025,
  title = {An {{Introduction}} to the {{Human Connectome Project}} for Early Psychosis},
  author = {Jacobs, Grace R. and Coleman, Michael J. and Lewandowski, Kathryn E. and Pasternak, Ofer and {Cetin-Karayumak}, Suheyla and {Mesholam-Gately}, Raquelle I. and Wojcik, Joanne and Kennedy, Leda and Knyazhanskaya, Evdokiya and Reid, Benjamin and Swago, Sophia and Lyons, Monica G. and Rizzoni, Elizabeth and John, Omar and Carrington, Holly and Kim, Nicholas and Kotler, Elana and Veale, Simone and Haidar, Anastasia and Prunier, Nicholas and Haaf, Moritz and Levitt, James J. and {Seitz-Holland}, Johanna and Rathi, Yogesh and Kubicki, Marek and Keshavan, Matcheri S. and Holt, Daphne J. and Seidman, Larry J. and {\"O}ng{\"u}r, Dost and Breier, Alan and Bouix, Sylvain and Shenton, Martha E.},
  year = 2025,
  month = may,
  journal = {Schizophrenia Bulletin},
  volume = {51},
  number = {3},
  pages = {658--671},
  issn = {1745-1701},
  doi = {10.1093/schbul/sbae123},
  langid = {english},
  pmcid = {PMC12061660},
  pmid = {39036958}
}

@article{taxCrossscannerCrossprotocolDiffusion2019a,
  title = {Cross-Scanner and Cross-Protocol Diffusion {{MRI}} Data Harmonisation: {{A}} Benchmark Database and Evaluation of Algorithms},
  shorttitle = {Cross-Scanner and Cross-Protocol Diffusion {{MRI}} Data Harmonisation},
  author = {Tax, Chantal MW. and Grussu, Francesco and Kaden, Enrico and Ning, Lipeng and Rudrapatna, Umesh and John Evans, C. and {St-Jean}, Samuel and Leemans, Alexander and Koppers, Simon and Merhof, Dorit and Ghosh, Aurobrata and Tanno, Ryutaro and Alexander, Daniel C. and Zappal{\`a}, Stefano and Charron, Cyril and Kusmia, Slawomir and Linden, David EJ. and Jones, Derek K. and Veraart, Jelle},
  year = 2019,
  month = jul,
  journal = {NeuroImage},
  volume = {195},
  pages = {285--299},
  issn = {1053-8119},
  doi = {10.1016/j.neuroimage.2019.01.077},
  urldate = {2024-07-22}
}

@article{vanessenWUMinnHumanConnectome2013c,
  title = {The {{WU-Minn Human Connectome Project}}: {{An}} Overview},
  shorttitle = {The {{WU-Minn Human Connectome Project}}},
  author = {Van Essen, David C. and Smith, Stephen M. and Barch, Deanna M. and Behrens, Timothy E.J. and Yacoub, Essa and Ugurbil, Kamil},
  year = 2013,
  month = oct,
  journal = {NeuroImage},
  volume = {80},
  pages = {62--79},
  issn = {10538119},
  doi = {10.1016/j.neuroimage.2013.05.041},
  urldate = {2025-12-09},
  langid = {english}
}

@article{ghodratiMRImageReconstruction2019,
  title = {{{MR}} Image Reconstruction Using Deep Learning: Evaluation of Network Structure and Loss Functions},
  shorttitle = {{{MR}} Image Reconstruction Using Deep Learning},
  author = {Ghodrati, Vahid and Shao, Jiaxin and Bydder, Mark and Zhou, Ziwu and Yin, Wotao and Nguyen, Kim-Lien and Yang, Yingli and Hu, Peng},
  year = 2019,
  month = sep,
  journal = {Quantitative Imaging in Medicine and Surgery},
  volume = {9},
  number = {9},
  pages = {1516--1527},
  issn = {2223-4292},
  doi = {10.21037/qims.2019.08.10},
  urldate = {2026-06-08},
  pmcid = {PMC6785508},
  pmid = {31667138}
}

@article{wasserthalTractSegFastAccurate2018a,
  title = {{{TractSeg}} - {{Fast}} and Accurate White Matter Tract Segmentation},
  author = {Wasserthal, Jakob and Neher, Peter and Maier-Hein, Klaus H.},
  date = {2018-12-01},
  journaltitle = {NeuroImage},
  shortjournal = {NeuroImage},
  volume = {183},
  pages = {239--253},
  issn = {1053-8119},
  url = {https://www.sciencedirect.com/science/article/pii/S1053811918306864},
  urldate = {2026-06-11}
}

@article{garyfallidisDipyLibraryAnalysis2014,
  title = {Dipy, a Library for the Analysis of Diffusion {{MRI}} Data},
  author = {Garyfallidis, Eleftherios and Brett, Matthew and Amirbekian, Bagrat and Rokem, Ariel and Van Der Walt, Stefan and Descoteaux, Maxime and Nimmo-Smith, Ian},
  date = {2014-02-21},
  journaltitle = {Frontiers in Neuroinformatics},
  shortjournal = {Front. Neuroinform.},
  volume = {8},
  publisher = {Frontiers},
  issn = {1662-5196},
  url = {https://www.frontiersin.org/journals/neuroinformatics/articles/10.3389/fninf.2014.00008/full},
  urldate = {2026-06-15},
  langid = {english}
}

@article{tournierMRtrix3FastFlexible2019a,
  title = {MRtrix3: {{A}} Fast, Flexible and Open Software Framework for Medical Image Processing and Visualisation},
  shorttitle = {MRtrix3},
  author = {Tournier, J-Donald and Smith, Robert and Raffelt, David and Tabbara, Rami and Dhollander, Thijs and Pietsch, Maximilian and Christiaens, Daan and Jeurissen, Ben and Yeh, Chun-Hung and Connelly, Alan},
  date = {2019-11-15},
  journaltitle = {NeuroImage},
  shortjournal = {NeuroImage},
  volume = {202},
  pages = {116137},
  issn = {1053-8119},
  url = {https://www.sciencedirect.com/science/article/pii/S1053811919307281},
  urldate = {2026-06-15}
}

@article{tournierRobustDeterminationFibre2007b,
  title = {Robust Determination of the Fibre Orientation Distribution in Diffusion {{MRI}}: {{Non-negativity}} Constrained Super-Resolved Spherical Deconvolution},
  shorttitle = {Robust Determination of the Fibre Orientation Distribution in Diffusion {{MRI}}},
  author = {Tournier, J-Donald and Calamante, Fernando and Connelly, Alan},
  date = {2007-05-01},
  journaltitle = {NeuroImage},
  shortjournal = {NeuroImage},
  volume = {35},
  number = {4},
  pages = {1459--1472},
  issn = {1053-8119},
  url = {https://www.sciencedirect.com/science/article/pii/S1053811907001243},
  urldate = {2026-06-15}
}

@inproceedings{ronnebergerUNetConvolutionalNetworks2015,
  title = {U-{{Net}}: {{Convolutional Networks}} for {{Biomedical Image Segmentation}}},
  shorttitle = {U-{{Net}}},
  booktitle = {Medical Image Computing and Computer-Assisted Intervention -- {{MICCAI}} 2015},
  author = {Ronneberger, Olaf and Fischer, Philipp and Brox, Thomas},
  editor = {Navab, Nassir and Hornegger, Joachim and Wells, William M. and Frangi, Alejandro F.},
  year = 2015,
  pages = {234--241},
  publisher = {Springer International Publishing},
  address = {Cham},
  doi = {10.1007/978-3-319-24574-4_28},
  isbn = {978-3-319-24574-4},
  langid = {english}
}

@article{haldarStateArtMR2026,
  title = {The “{{State}} of the {{Art}}” in {{MR Image Reconstruction}}? {{Knowledge}}, {{Culture}}, and {{What We Leave Behind}} in {{An Era}} of {{Big Data}} and {{Machine Learning}}},
  shorttitle = {The “{{State}} of the {{Art}}” in {{MR Image Reconstruction}}?},
  author = {Haldar, Justin P.},
  year = {2026},
  journaltitle = {Magnetic Resonance in Medicine},
  shortjournal = {Magn. Reson. Med.},
  volume = {96},
  number = {1},
  pages = {7--12},
  issn = {1522-2594},
  url = {https://onlinelibrary.wiley.com/doi/abs/10.1002/mrm.70377},
  urldate = {2026-07-15},
  langid = {english}
}

\end{document}